\documentclass[10pt,twocolumn,letterpaper]{article}

\usepackage{iccv}   

\usepackage{multirow}
\usepackage{amsmath}
\usepackage{mathtools}
\usepackage{amsthm}
\usepackage{adjustbox}
\definecolor{row_gray}{rgb}{0.95,0.95,0.95}
\definecolor{tab_blue}{rgb}{0.21, 0.50, 0.75}
\definecolor{tab_red}{rgb}{0.7,0.2,0.2}
\definecolor{iccv_blue}{rgb}{0.21,0.49,0.74}
\usepackage[pagebackref,breaklinks,colorlinks,allcolors=iccv_blue]{hyperref}

\title{One-Step Diffusion Model for Image Motion-Deblurring}

\author{
    Xiaoyang Liu$^{1}$, \enspace
    Yuquan Wang$^{1}$, \enspace
    Zheng Chen$^{1}$, \\
    Jiezhang Cao$^{2}$, \enspace
    He Zhang$^{3}$, \enspace
    Yulun Zhang$^{1}$\thanks{Corresponding author: Yulun Zhang, yulun100@gmail.com}, \enspace
    Xiaokang Yang$^{1}$ \\
  \textsuperscript{1}Shanghai Jiao Tong University,\enspace
  \textsuperscript{2}Harvard University,\enspace
  \textsuperscript{3}Adobe Research\enspace 
  \vspace{-4.mm}
}

\begin{document}
\maketitle

\begin{abstract}
Currently, methods for single-image deblurring based on CNNs and transformers have demonstrated promising performance. However, these methods often suffer from perceptual limitations, poor generalization ability, and struggle with heavy or complex blur. While diffusion-based methods can partially address these shortcomings, their multi-step denoising process limits their practical usage. In this paper, we conduct an in-depth exploration of diffusion models in deblurring and propose a \textbf{o}ne-\textbf{s}tep \textbf{d}iffusion model for \textbf{d}eblurring (OSDD), a novel framework that reduces the denoising process to a single step, significantly improving inference efficiency while maintaining high fidelity. To tackle fidelity loss in diffusion models, we introduce an enhanced variational autoencoder (eVAE), which improves structural restoration. Additionally, we construct a high-quality synthetic deblurring dataset to mitigate perceptual collapse and design a dynamic dual-adapter (DDA) to enhance perceptual quality while preserving fidelity. Extensive experiments demonstrate that our method achieves strong performance on both full and no-reference metrics. Our code and pre-trained model will be publicly available at \url{https://github.com/xyLiu339/OSDD}.

\end{abstract}

\setlength{\abovedisplayskip}{2pt}
\setlength{\belowdisplayskip}{2pt}

\vspace{-1mm}
\section{Introduction}
\label{sec:intro}
\vspace{-1mm}

Image deblurring, a key sub-task of image restoration, focuses on recovering sharp images from blurry ones. The blurring artifacts may result from various factors such as camera shake, fast-moving objects, or lens imperfections~\cite{zhang2022deep}. Restoring a high-quality image from its degraded version is a highly challenging problem due to the ill-posed nature of the task, which means there are multiple possible solutions for any given blurred image.

\begin{figure}[t]
\scriptsize
\centering
\begin{tabular}{cc}
\hspace{-0.45cm}
\begin{adjustbox}{valign=t}
\begin{tabular}{c}
\includegraphics[width=0.1420\textwidth,height=0.183\textwidth]{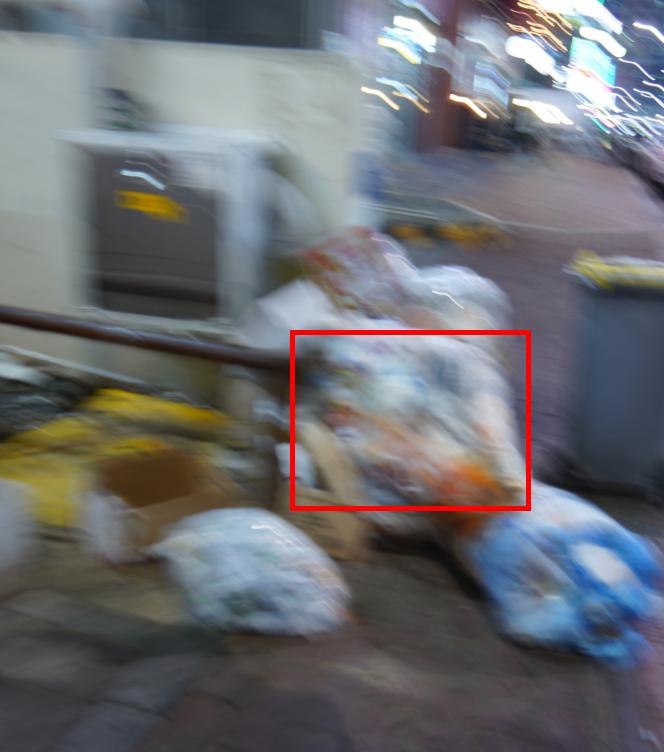}
\\
scene108-20
\end{tabular}
\end{adjustbox}
\hspace{-4.7mm}
\begin{adjustbox}{valign=t}
\begin{tabular}{cccccc}
\includegraphics[width=0.108\textwidth]{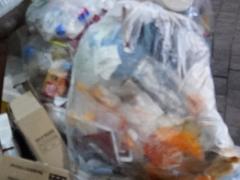} \hspace{-4.5mm} &
\includegraphics[width=0.108\textwidth]{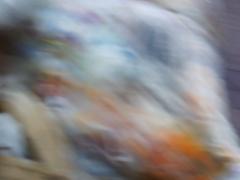} \hspace{-4.5mm} &
\includegraphics[width=0.108\textwidth]{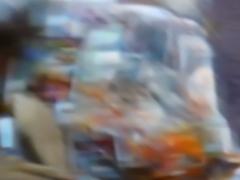} \hspace{-4.5mm} 
\\
HQ \hspace{-4.5mm}
 &
Blurry \hspace{-4.5mm}  &
AdaRevD~\cite{xintm2024AdaRevD} \hspace{-4.5mm} &
\\
\includegraphics[width=0.108\textwidth]{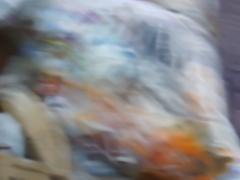} \hspace{-4.5mm} &
\includegraphics[width=0.108\textwidth]{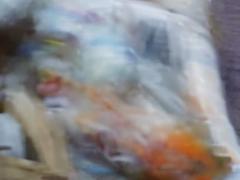} \hspace{-4.5mm} &
\includegraphics[width=0.108\textwidth]{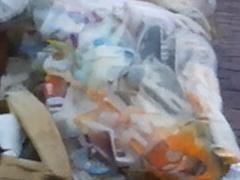} \hspace{-4.5mm}  
\\ 
HI-Diff~\cite{hi-diff} \hspace{-4.5mm}  &
OSEDiff~\cite{wu2024one}\hspace{-4.5mm} &
OSDD (ours) \hspace{-4.5mm}
\\
\end{tabular}
\end{adjustbox}

\end{tabular}
\vspace{-3.5mm}
\caption{Visual debluring comparison on RealBlur-J. Our model shows robust capabilities of handling complex blurs and restoring clear boundaries, providing a better visual experience.}
\label{fig:first_comp}
\vspace{-6.5mm}
\end{figure}

With the rapid advancement of CNNs and transformers, significant progress has been made in image deblurring. CNNs, with their ability to learn hierarchical representations from data, have greatly improved the accuracy and efficiency of image deblurring~\cite{Chakrabarti2016a, Tao2018scale}. Meanwhile, transformers, which excel at capturing long-range dependencies and contextual information, have further propelled the evolution of deblurring techniques~\cite{Zamir2021restormer, Tsai2022Stripformer}, particularly in handling dynamic blur patterns. Despite their promising deblurring performance, these methods often suffer from perceptual limitations (\cref{fig:first_comp}), with generated deblurred images showing noticeable differences from real-world images. Additionally, their models exhibit limited generalization, struggling with unseen or complex blur patterns.

Recently, diffusion models~\cite{ldm, sd3} have attracted increasing attention due to their powerful generative capabilities. Applying diffusion models to deblurring can alleviate some of the aforementioned issues, benefiting from large-scale pretraining on high-quality datasets. However, the multi-step denoising process consumes substantial time, hindering practical application. In this work, we address these limitations by proposing a \textbf{o}ne-\textbf{s}tep \textbf{d}iffusion model for \textbf{d}eblurring (\textbf{OSDD}), which reduces the denoising process to a single step, inspired by \cite{li2024distillation, wu2024one, wang2024sinsr}.

However, the diffusion models face two key challenges. The first is that image-to-image diffusion models inherently involve a \textbf{trade-off} between \textit{fidelity} and \textit{perceptual quality}. Specifically, due to inherent structural bottlenecks, they often struggle to preserve fine details from input images and may generate artifacts that were not present in the original image, which limits their effectiveness in high-fidelity tasks (\eg, deblurring). 
To address this, we introduce an \textbf{Enhanced Variational Autoencoder (eVAE)} that mitigates the fidelity loss caused by latent space compression in diffusion models, which significantly improves both pixel-level and structural-level restoration without requiring fine-tuning of the latent diffusion model.

Another major challenge in the diffusion model is the domain gap between image generation and restoration, \ie, diffusion models can barely handle blurry latents, making it difficult to leverage large-scale pre-training and achieve effective transfer learning. Fine-tuning on existing deblurring datasets~\cite{Nah2017deep, Rim2020real} leads to a severe decline in perceptual quality—a phenomenon we term \textbf{perceptual collapse}, which goes beyond the conventional fidelity-perception trade-off. To solve this, we introduce a \textbf{high-quality synthetic deblurring dataset} for training diffusion models. This approach leverages the pre-training capabilities of diffusion models, enabling them to learn the deblurring task while generating realistic, high-quality images.

To further leverage the results of transfer learning with our synthetic dataset and enable the model to learn real-world blur, we propose a \textbf{D}ynamic \textbf{D}ual-\textbf{A}dapter \textbf{(DDA)} mechanism, which dynamically fuses the two pre-trained models during inference using a manually adjustable weight factor $\gamma$. This design allows for flexible control over the learning impact between the existing datasets and our high-quality synthetic datasets, effectively preventing perceptual collapse while maintaining high fidelity.

Extensive experiments show that our eVAE with the high-quality synthetic data pre-training and our carefully-designed DDA, achieves state-of-the-art performance on both full-reference metrics (\eg, LPIPS~\cite{lpips}, DISTS~\cite{dists}) and no-reference metrics (\eg, NIQE~\cite{niqe}, MUSIQ~\cite{musiq}). Our main contribution can be summarized as follows:
\begin{itemize}
    \item We introduce a one-step diffusion framework, OSDD, for single-image deblurring, achieving competitive performance while drastically reducing inference time.
    \item We identify and address the structural limitations of diffusion models for deblurring by proposing an enhanced VAE (eVAE) and training strategy, improving fidelity.
    \item We construct a high-quality synthetic deblurring dataset using synthetic blur kernels, enabling effective transfer learning and significantly enhancing the perceptual quality of diffusion-based deblurring models.
    \item We design a novel adapter mechanism that dynamically improves perceptual quality while preserving fidelity during inference, enabling high-quality deblurring.
\end{itemize}

\vspace{-2mm}
\section{Related Works}
\vspace{-1.5mm}
\subsection{Image Deblurring}
\vspace{-1.5mm}

\noindent \textbf{Traditional Methods and Hand-crafted Priors.}
Traditional image deblurring methods have typically relied on hand-crafted priors to constrain the solution space, especially in the early stages of image deblurring research~\cite{10.1145/1179352.1141956, 10.1145/1360612.1360672, 5206815}. These methods are based on assumptions such as sparsity in image gradients~\cite{Xu2013unnatural}, intensity or color sparsity~\cite{nimisha2017blur}, and internal patch recurrence~\cite{michaeli2014blind}, which help in removing blur. However, these hand-crafted priors often fail to exploit the deeper structure of the image, making the optimization process complex. Moreover, traditional methods often rely on simpler assumptions, such as uniform blur~\cite{xu2010two, Xu2013unnatural} across the image, which limits their performance when dealing with more complex scenes (\eg, dynamic objects and non-uniform blur).

\noindent \textbf{Deep Learning Methods for Kernel Estimation.}
With the rise of deep learning, CNN-based methods for blur kernel estimation have become more common in image deblurring. Early approaches estimated the blur kernel and used traditional restoration methods for deblurring~\cite{Chakrabarti2016a, Sun2015learning, Schuler2016learning}. While effective, inaccuracies in kernel estimation often led to visible artifacts in the deblurred images.

\noindent \textbf{End-to-End Deblurring.}
In recent years, end-to-end deblurring methods powered by CNNs and transformers have significantly advanced the field of image deblurring. CNNs are widely used to directly restore sharp images from blurry ones, eliminating the need for explicit kernel estimation. Approaches~\cite{Tao2018scale, Abuolaim2020defocus, Zhang2019DMPHN} enhance detail recovery through residual learning~\cite{pan2018learning}, skip connections~\cite{Gao2019dynamic}, and multi-scale strategies~\cite{Nah2017deep}, achieving remarkable progress. Simultaneously, the introduction of transformers has provided a new way to model global context, with many recent studies~\cite{Wang2022uformer, Zamir2021restormer, Tsai2022Stripformer} utilizing the self-attention mechanism of transformers to capture long-range dependencies.

\vspace{-0.5mm}
\subsection{Diffusion Models}
\vspace{-0.5mm}
Diffusion Models (DMs)~\cite{ddpm, ldm, song2020denoising, sd3} are probabilistic generative models that synthesize data from Gaussian noise. DMs have demonstrated remarkable performance in super-resolution~\cite{li2024distillation, wu2024one, wang2024sinsr} and various image restoration tasks, such as image inpainting~\cite{ldm} and deblurring~\cite{10446822, 10483663, Spetlik_2024_WACV}. DiffIR~\cite{Xia_2023_ICCV} and HI-Diff~\cite{hi-diff} leverage diffusion models to generate prior representations and adopt a two-stage training strategy. DvSR~\cite{Whang_2022_CVPR} and Ren \etal~\cite{Ren_2023_ICCV} introduce conditional diffusion models into image deblurring tasks, integrating residual models and multiscale information. ID-Blau~\cite{Wu_2024_CVPR} proposes a blur augmentation method based on implicit diffusion, generating diverse blurred images under controllable conditions, which enhances performance when integrated into other deblurring models.

However, DMs require a large number of iterative steps and operate in the image space to ensure high-quality reconstruction, resulting in substantial computational costs. Additionally, diffusion-based models tend to overlook fine image details, which leads to suboptimal performance.

\vspace{-0.5mm}
\subsection{Deblurring Datasets}
\vspace{-0.5mm}
In motion deblurring, several approaches have been proposed to generate blurred training datasets, each with its advantages and limitations. Video-based datasets, created by averaging consecutive frames from high-speed video sequences~\cite{Shen2019human, Nah2017deep}, efficiently generate blur but are limited by specific scenes and camera settings, reducing generalizability. Real-world datasets~\cite{Rim2020real} offer high authenticity and better generalization to natural motion blur but are time-consuming to collect and often have limited sample sizes. Synthetic blurring~\cite{Sun2015learning, NIPS2014_1c1d4df5, Kupyn2018deblurgan} using predefined kernels allows precise control over blur types but may lack the complexity of real-world blur, limiting generalization. Each method offers trade-offs between realism, scalability, and diversity, and the choice of approach depends on the specific objectives and constraints of the study.

\vspace{-1.5mm}
\section{Methodology}
\vspace{-1mm}
Single-image blind deblurring aims to restore a sharp high-quality image $x_h$ from a blurry image $x_l$. While diffusion models~\cite{ldm, sd3} can address the generalization and generation quality issues of CNNs and transformers, the multi-step denoising process is time-consuming. To overcome this, we propose a one-step diffusion model for deblurring in~\cref{sec: ldm} with appropriate modifications.

We identify a significant bottleneck in the model when prioritizing fidelity. To address this, we introduce an enhanced VAE (eVAE) architecture, as discussed in~\cref{sec: vae}.
Besides, we observe poor visual quality and suboptimal performance on no-reference metrics in the deblurring results, a phenomenon we term perceptual collapse. To mitigate this, we construct a high-quality synthetic deblurring dataset using random synthetic kernels (will be discussed in~\cref{sec: data}). Accordingly, we design a dynamic dual-adapter (DDA) to effectively merge the trained models from the synthetic and real-world datasets (\eg, GoPro~\cite{Nah2017deep} and RealBlur~\cite{Rim2020real}), as detailed in~\cref{sec: dda}.

\vspace{-1mm}
\subsection{VAE}
\label{sec: vae}
\vspace{-1mm}
Variational Autoencoders (VAEs) are widely used for data generation, dimensionality reduction, and representation learning. They learn a latent space that captures data structure and allows for smooth interpolations. In latent diffusion models (LDMs), VAEs encode high-dimensional images into a compact latent space, improving computational efficiency and image quality. This reduces the complexity of diffusion, enabling high-resolution image synthesis with fewer resources and better semantic control for tasks like text-to-image generation and image inpainting.

\begin{figure}[t]
\begin{center}
\centerline{\includegraphics[width=\linewidth]{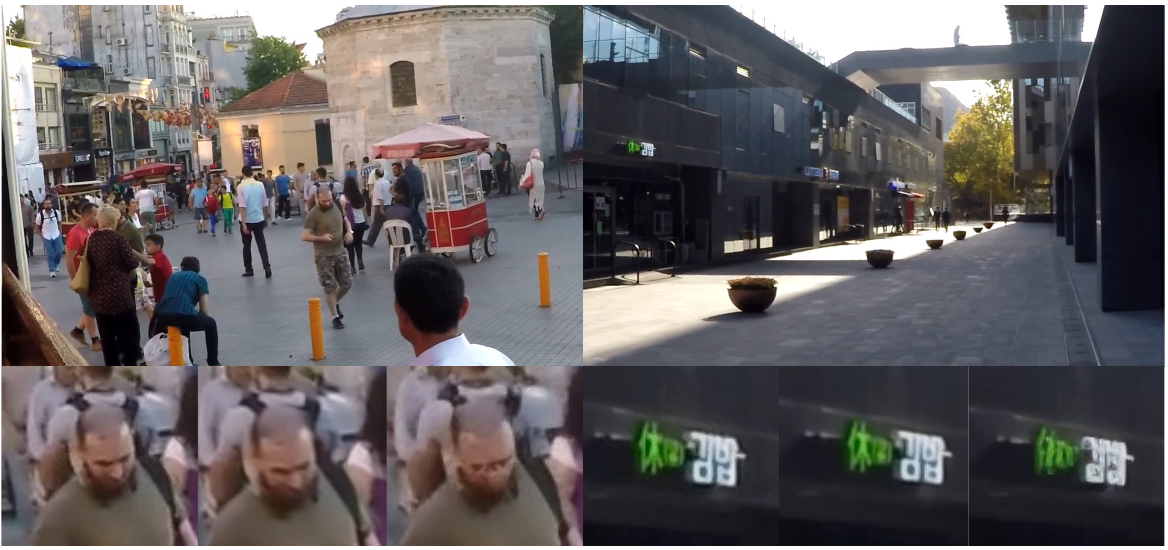}}
\vspace{-2.5mm}
\caption{Some samples from the GoPro~\cite{Nah2017deep} test set are shown. At the bottom of each sample, from left to right, we present the high-quality (HQ) image, AdaRevD-L~\cite{xintm2024AdaRevD}, and VAE reconstruction. It is evident that the VAE even fails to accurately reconstruct the HQ image in the deblurring dataset, which significantly limits the upper bound of the entire diffusion model's performance.}
\vspace{-12mm}
\label{fig: vae_only}
\end{center}
\end{figure}

\vspace{-1mm}
\subsubsection{Limitation of VAE}
\vspace{-0.5mm}
The effectiveness of VAEs diminishes in tasks that require strict conditioning, such as image restoration, where maintaining fidelity to the input image is crucial. A common limitation observed in these tasks is the structural and pixel-level discrepancies in the generated output, showing noticeable deviations from the original input image. \cref{fig: vae_only} illustrates this point. Even when using the VAE to directly encode and decode the high-quality (HQ) image, the reconstructions show substantial distortions. This limitation makes it more difficult to effectively incorporate VAE into deblurring latent diffusion models.

This issue primarily stems from the downsampling factor \( f \) and the dimensionality \( c \) of the latent space. VAE compresses an input image of size \( H \times W \times 3 \) into a lower-dimensional representation of \( h \times w \times c \), where \( h = H / f \), \( w = W / f \). For instance, in Stable Diffusion 2.1~\cite{ldm}, the default setting adopts \( f = 8 \) and \( c = 4 \), meaning that each image undergoes a substantial reduction in spatial resolution before being processed in the latent space. That considerable amount of fine detail is irreversibly lost during encoding, making perfect reconstruction infeasible. This inherent loss of information becomes more pronounced as image complexity increases, resulting in significant discrepancies when attempting to restore fidelity.

\subsubsection{Enhanced VAE}
\begin{figure*}[t]
\begin{center}
\centerline{\includegraphics[width=1\linewidth]{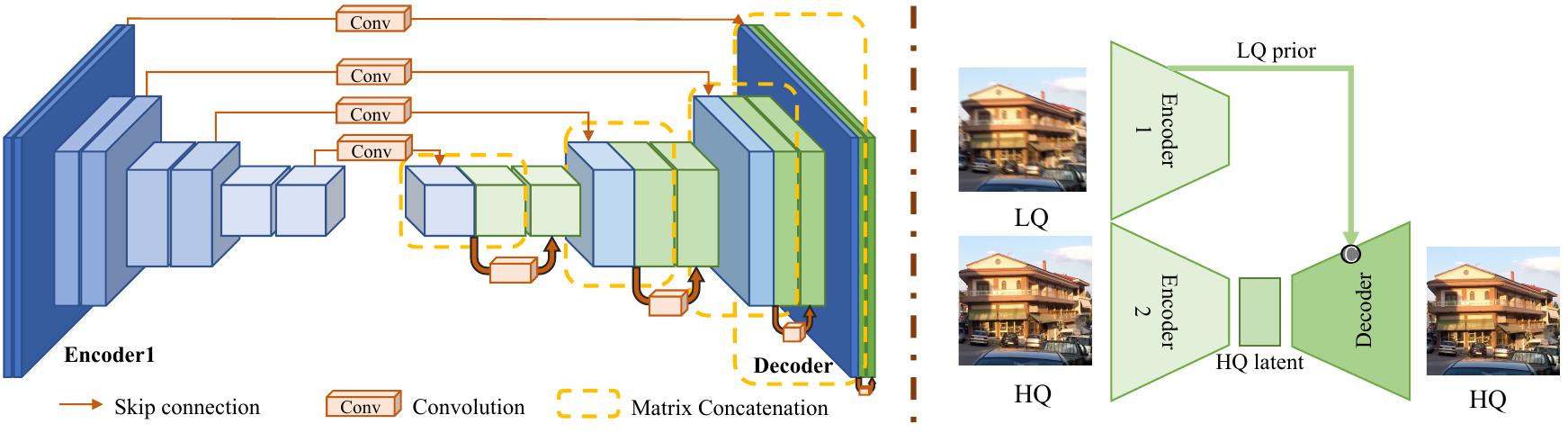}}
\vspace{-3mm}
\caption{On the left, we illustrate the design of our enhanced VAE, where features from the encoder are concatenated to the decoder through a convolutional block. On the right, we present the training strategy of the eVAE. Encoder 1 handles the skip-connection features, while Encoder 2 is responsible for encoding the high-quality image.}
\vspace{-12mm}
\label{fig: vae}
\end{center}
\end{figure*}

As required in the field of image restoration, the generated image should adhere more closely to the original input, preserving as much detail as possible. However, conventional VAEs struggle to achieve this due to their inherent information loss. To address this issue, we propose a simple yet effective enhanced VAE (eVAE), inspired by the skip connections in U-Net. Specifically, we introduce intermediate connections between the corresponding layers of the encoder and decoder, enabling the decoder to leverage more high-frequency and image-adjacent information to improve reconstruction quality, details in~\cref{fig: vae}.

In our design, for each resolution level in the encoder, we extract features from the last layer before downsampling, pass them through some convolutional layers in the skip connection, and concatenate them with the corresponding first layer in the decoder. This mechanism helps the model retain fine-grained details, improving image reconstruction quality and adherence to the original input.

\subsubsection{Training of eVAE}
\label{sec: vaeft}
To better adapt our eVAE to the deblurring dataset and enable it to capture high-frequency details, we employ two encoders and a single decoder, all trained simultaneously. As illustrated in~\cref{fig: vae}, one encoder ($E_1$) takes the blurry image $x_l$ as input and provides skip connection features for the decoder, while the other encoder ($E_2$) takes the sharp image $x_h$ as input to supply a high-quality latent for the decoder. The decoder receives both the representation from the sharp image and the skip connection features. The rationale behind this is that during the inference phase, the latent $\hat{z}_h$ obtained by encoding and latent deblurring of $x_l$ should be almost identical to the latent $z_h$ of $x_h$.

During training, the loss function is defined as  
\begin{equation}
\label{vaeloss}
    \mathcal{L} = \mathcal{L}_1 + \alpha \cdot \mathcal{L}_{p}.
\end{equation}
To prevent the model from learning overly smoothed representations, the weight \(\alpha\) of the perception loss is adjusted based on the scale of the two.

\vspace{-1mm}
\subsection{Latent Diffusion}
\label{sec: ldm}
\vspace{-0.5mm}
\subsubsection{Preliminary}
\vspace{-0.5mm}
In the standard diffusion process, we gradually add Gaussian noise over \( T \) steps, following a Markov chain:  
\begin{align}
    & q(z_{1:T} | z_0) = \prod_{t=1}^{T} q(z_t | z_{t-1}), \\
    & q(z_t | z_{t-1}) = \mathcal{N} (z_t; \sqrt{1 - \beta_t} z_{t-1}, \beta_t I),
\end{align}
where \( \beta_t \in (0,1) \) controls the variance of the noise, and \( \mathcal{N}(\cdot; \mu, \Sigma) \) denotes a Gaussian distribution with mean \( \mu \) and covariance \( \Sigma \). By reparameterization, the noisy sample at step \( t \) can be directly expressed as:  
\begin{equation}
    q(z_t | z_0) = \mathcal{N}(z_t; \sqrt{\bar{\alpha}_t} z_0, (1 - \bar{\alpha}_t)I),
\end{equation}
where \( \alpha_t = 1 - \beta_t \) and \( \bar{\alpha}_t = \prod_{i=1}^{t} \alpha_i \).  

In the reverse process, we reconstruct \( z_0 \) from a Gaussian prior \( z_T \) through a Markov chain running backward: 
\begin{equation}
    q(z_{t-1} | z_t, z_0) = \mathcal{N} (z_{t-1}; \mu_t(z_t, z_0), \sigma_t^2 I),
\end{equation}
where the mean \( \mu_t(z_t, z_0) \) is given by:
\begin{equation}
    \mu_t(z_t, z_0) = \frac{1}{\sqrt{\alpha_t}} \left( z_t - \frac{1 - \alpha_t}{\sqrt{1 - \bar{\alpha}_t}} \boldsymbol{\epsilon} \right).
\end{equation}
Here, \( \boldsymbol{\epsilon} \) represents the noise in \( z_t \). To estimate \( \boldsymbol{\epsilon} \), a denoising network \( \epsilon_\theta \) predicts the noise at each step conditioned on the latent feature \( z_t \) and a text \( c \), denoted as $\boldsymbol{\epsilon} = \epsilon_\theta (z_t, c, t)$.

Substituting \( \epsilon_\theta \) into the reverse step equation and setting the variance to \( (1 - \alpha_t) \), the update rule becomes:
\begin{equation}
    z_{t-1} = \frac{1}{\sqrt{\alpha_t}} \left( z_t - \frac{1 - \alpha_t}{\sqrt{1 - \bar{\alpha}_t}} \epsilon_{\theta}(z_t, c, t) \right) + \sigma_t \epsilon_t,
\end{equation}
where \( \epsilon_t \sim \mathcal{N}(0, I) \). Besides, the research~\cite{song2020denoising} shows that modeling the diffusion process as a non-Markov chain eliminates the need for step-by-step denoising, significantly improving inference speed.
\begin{figure*}[ht]
\vspace{-5pt}
\begin{center}
\centerline{\includegraphics[width=1\linewidth]{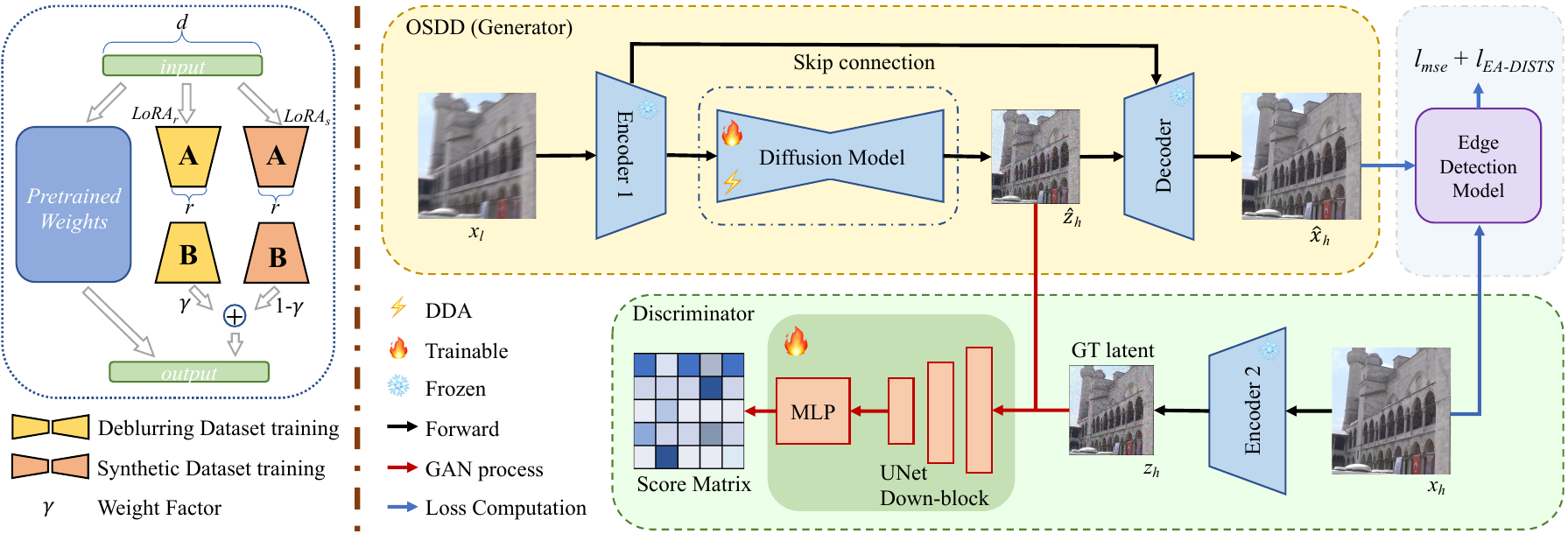}}
\vspace{-3mm}
\caption{On the left, the specific structure and operational principles of the Dynamic Dual-Adapter (DDA) are illustrated. On the right, we show the pipeline of our latent-diffusion model training. The frozen VAE modules are inherited from the first-stage training(\cref{sec: vaeft}).}
\vspace{-13mm}
\label{fig: pipeline}
\end{center}
\end{figure*}

\begin{figure}[t]
\begin{center}
\vspace{4pt}
\centerline{\includegraphics[width=0.98\linewidth]{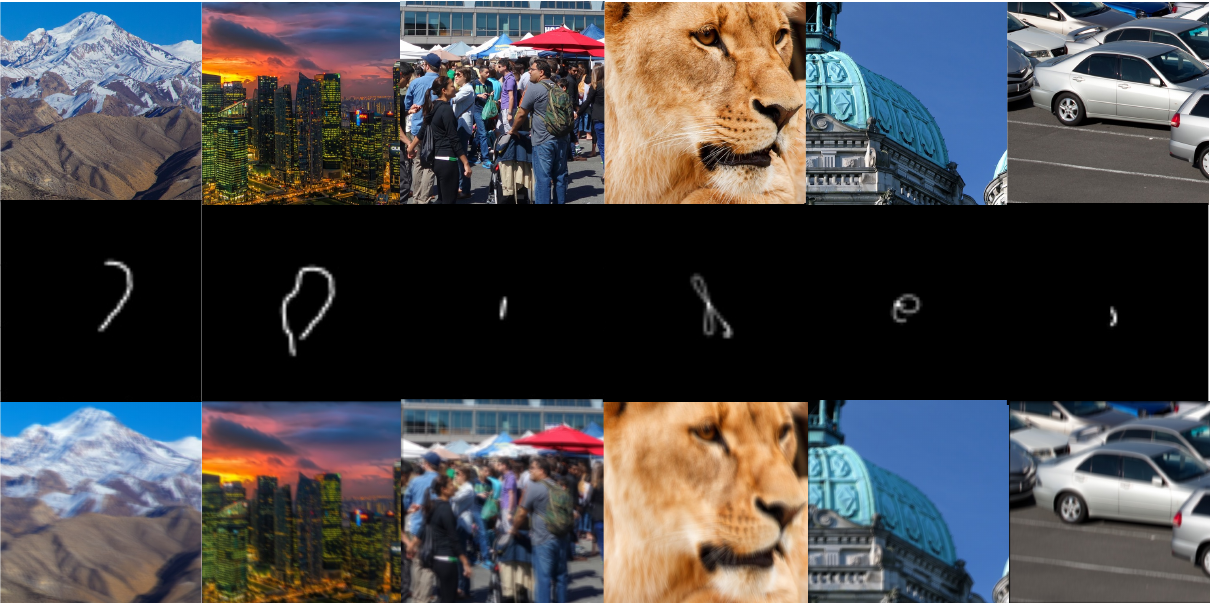}}
\vspace{-3mm}
\caption{Examples of some randomly generated kernel trajectories and corresponding LQ and HQ images.}
\vspace{-14mm}
\label{fig: syn}
\end{center}
\end{figure}

\subsubsection{One-Step Diffusion Model for Deblurring}
With the recent advancements in diffusion models, researchers have explored various methods to gradually reduce the number of inference steps, successfully shortening to a single step. This approach has been extensively investigated in tasks such as super-resolution~\cite{li2024distillation, wu2024one, wang2024sinsr}.

In this work, we extend the single-step diffusion model to the task of image deblurring and propose the one-step diffusion model for deblurring (OSDD). The whole pipeline is shown in~\cref{fig: pipeline} (right part). Specifically, we first employ the encoder of our eVAE to compress the low-quality image \( x_l \) into the latent space, obtaining \( z_l \), which corresponds to \( z_t \). Subsequently, we directly predict the clean latent variable \( \hat{z}_h \) using the model's estimated noise:
\begin{equation}
    \hat{z}_h = \frac{z_t - \sqrt{1 - \bar{\alpha}_t} \hat{\epsilon}}{\sqrt{\bar{\alpha}_t}},\quad  \hat{\epsilon} = \epsilon_\theta (z_t, c, t) .  
\end{equation}

There has been extensive exploration of c. To accommodate diverse blurry image inputs, we adopt a lightweight, learnable text embedding, aiming for this embedding to serve as an empirical deblurring prompt that guides the deblurring process. For \( t \), we select a fixed value that remains unchanged during both training and testing. 

Additionally, we employ a latent discriminator, inspired by~\cite{sauer2024fast, sauer2025adversarial, li2024distillation} to constrain the model’s convergence process. The discriminator incorporates restored latent $\hat{z}_h$, together with the $z_h$ encoded from the high-quality image $x_h$ by Encoder 2. The discriminator consists of a U-Net downblock, followed by a multi-layer perceptron (MLP), to distinguish between real and fake samples.

\vspace{-1mm}
\subsubsection{Dynamic Dual-Adapter}
\label{sec: dda}
\vspace{-1mm}
Our OSDD is designed to learn from both high-quality synthetic data and real-world blurred data. Achieving high-quality restoration while addressing real-world blur scenarios is crucial. To this end, we propose a Dynamic Dual-Adapter (DDA) that enhances the model's deblurring performance effectively. As shown in~\cref{fig: pipeline} on the left, we employ two distinct sets of LoRA~\cite{lora} modules $LoRA_r$ and $LoRA_s$, one fine-tuned on datasets like~\cite{Nah2017deep, Rim2020real} while another on our high-quality synthetic dataset (\cref{sec: data}), allowing the model to simultaneously learn real-world blur characteristics while maintaining high generative quality. During the inference phase, both trained LoRA modules are loaded, and their influence is dynamically balanced by adjusting their weight factors $\gamma$:
\begin{equation}
\label{eq: lora}
    y_o = W(y_i) + \gamma LoRA_r(y_i) + (1-\gamma) LoRA_s(y_i),
\end{equation}
where $y_i, y_o$ denote the input and output of each linear or convolutional layer that requires updating, respectively. W represents the original weight matrix of the layer. Each LoRA layer consists of two low-rank matrices. Specifically, when $\gamma=1$, we use the weight trained on~\cite{Nah2017deep, Rim2020real}, while decreasing $\gamma$ gives more weight to the model trained on our synthetic dataset. This adaptive strategy integrates the strengths of both modules, ensuring effective blur removal while significantly enhancing perceptual quality.

\subsubsection{Training of Diffusion Model}
\label{sec: ldmtrain}
During training, the loss function of our diffusion model is composed of the following terms:
\begin{align}
\label{ldmloss}
    &\mathcal{L} = \mathcal{L}_2(\hat{x}_h, x_h) + \lambda_1 \mathcal{L}_{\textit{EA\_DISTS}}(\hat{x}_h, x_h) + \lambda_2 \mathcal{L}_{G}(\hat{z}_h), \\
    &\mathcal{L}_{G} = -\mathbb{E}_{x_l}(\log \mathcal{D}(\hat{z}_h)), \\
    &\mathcal{L}_{D} = -\mathbb{E}_{x_h}(\log \mathcal{D}(z_h)) -\mathbb{E}_{x_l}(\log (1-\mathcal{D}(\hat{z}_h))),
\end{align}
where \(\mathcal{L}_{\textit{EA\_DISTS}}\) is the edge-aware DISTS loss~\cite{li2024distillation}, which outperforms traditional perceptual loss (\eg, LPIPS~\cite{lpips}), and \(\mathcal{L}_D\) denotes the loss of the latent discriminator.

\begin{table*}[!ht]
    \centering
    \setlength{\tabcolsep}{1mm}
    \resizebox{1.0\textwidth}{!}{
    \begin{tabular}{l|cccc|cccc|cccc} \toprule
       \rowcolor{row_gray} & \multicolumn{4}{c|}{GoPro} & \multicolumn{4}{c|}{RealBlur-J} & \multicolumn{4}{c}{RealBlur-R} \\ 
       \rowcolor{row_gray} \multirow{-2}{*}{Method} & CLIPIQA$ \uparrow $ & NIQE$ \downarrow $ & MUSIQ$ \uparrow $ & MANIQA$ \uparrow $ & CLIPIQA$ \uparrow $ & NIQE$ \downarrow $ & MUSIQ$ \uparrow $ & MANIQA$ \uparrow $ & CLIPIQA$ \uparrow $ & NIQE$ \downarrow $ & MUSIQ$ \uparrow $ & MANIQA$ \uparrow $ \\ 
       \midrule
       MPRNet~\cite{Zamir2021multi} & 0.2537 & 5.16 & 44.18 & 0.5230 & 0.2281 & 5.31 & 48.44 & 0.5545 & 0.2021 & 8.24 & 30.27 & 0.5052 \\
       HINet~\cite{Chen2021hinet} & 0.2513 & 5.08 & 44.07 & 0.5227 & 0.2176 & 5.07 & 44.96 & 0.5362 & 0.2036 & 8.06 & 29.54 & 0.5057 \\
       Restormer~\cite{Zamir2021restormer} & 0.2557 & 5.18 & 44.96 & 0.5269 & 0.2323 & 5.23 & 48.43 & 0.5639 & 0.1984 & 7.88 & 31.08 & 0.5224 \\
       Uformer~\cite{Wang2022uformer} & 0.2588 & 5.18 & 44.81 & 0.5265 & 0.2415 & 5.12 & 50.30 & 0.5809 & 0.2088 & 7.94 & 31.91 & \textcolor{tab_blue}{0.5296} \\
       NAFNet~\cite{Chen2022simple} & \textcolor{tab_red}{0.2603} & 5.10 & 45.28 & 0.5355 & 0.2248 & 5.05 & 47.07 & 0.5514 & \textcolor{tab_blue}{0.2117} & 7.96 & 31.44 & 0.5257 \\
       Stripformer~\cite{Tsai2022Stripformer} & 0.2444 & 4.98 & \textcolor{tab_blue}{46.00} & 0.5345 & N/A & N/A &N/A & N/A & N/A& N/A & N/A & N/A \\
       UFPNet~\cite{fang2023UFPNet} & 0.2573 & 5.11 & 45.33 & 0.5364 & \textcolor{tab_blue}{0.2658} & 5.45 & 51.31 & 0.5797 & \textcolor{tab_red}{0.2193} & 7.84 & 31.02 & 0.5264 \\
       AdaRevD~\cite{xintm2024AdaRevD} & 0.2600 & 5.03 & 45.79 & \textcolor{tab_blue}{0.5440} & 0.2469 & 5.22 & 51.34 & 0.5802 & 0.2123 & 7.91 & 31.27 & 0.5239 \\ 
       
       \midrule
       HI-Diff~\cite{hi-diff}  & 0.2584 & 5.19 & 45.55 & 0.5360 & 0.2430 & 5.24 & 50.55 & 0.5771 & 0.2001 & 7.94 & 31.48 & 0.5243 \\
       DiffIR~\cite{Xia_2023_ICCV} & \textcolor{tab_red}{0.2603} & 5.16 & 45.67 & 0.5388 & 0.2433 & 5.20 & 48.79 & 0.5639 & 0.2017 & 7.94 & 31.40 & 0.5224 \\
       OSEDiff~\cite{wu2024one} & 0.2123 & 4.48 & 39.92 & 0.4859 & 0.2178 & 5.10 & 41.10 & 0.5105 & 0.1994 & 8.08 & 28.89 & 0.5041 \\ \midrule
       OSDD ($\gamma=1$) & 0.2067 & \textcolor{tab_blue}{4.23} & 44.58 & 0.5114 & 0.2435 & \textcolor{tab_blue}{4.57} & \textcolor{tab_blue}{53.40} & \textcolor{tab_blue}{0.5828} & 0.2027 & \textcolor{tab_blue}{7.24} & \textcolor{tab_blue}{32.64} & 0.5249 \\ 
       OSDD ($\gamma=0.7$) & 0.2421 & \textcolor{tab_red}{4.11} & \textcolor{tab_red}{51.19} & \textcolor{tab_red}{0.5536} & \textcolor{tab_red}{0.2720} & \textcolor{tab_red}{4.54} & \textcolor{tab_red}{55.08} & \textcolor{tab_red}{0.6035} & 0.1980 & \textcolor{tab_red}{6.78} & \textcolor{tab_red}{32.66} & \textcolor{tab_red}{0.5353} \\ \bottomrule
    \end{tabular}}
    \vspace{-3.mm}
    \caption{No-reference image quality metrics are adopted here. Our model pre-trained on our synthetic dataset can get excellent performance on both GoPro and RealBlur datasets compared to other baselines.}
    \vspace{-5.mm}
    \label{tab: iqa_4}
\end{table*}

\vspace{-2mm}
\section{Synthetic Data Generation}
\label{sec: data}
\vspace{-1.5mm}
Directly training on deblurring datasets can lead to perceptual collapse, as we will demonstrate in~\cref{sec: experiments} and~\cref{tab: iqa_4}. This issue primarily arises due to the limited scale of contemporary deblurring datasets (compared to the training sets used for diffusion models), their lack of diversity in scenes, and their relatively low quality (attributed to outdated recording equipment and technological limitations of the time).
Considering the inherent misalignment between diffusion models and the deblurring task, as well as to fully leverage the pre-trained capabilities of diffusion models, we attempt to fine-tune the diffusion model by synthesizing blurry-sharp image pairs through manually applied blur kernels on real images. Specifically, we adopt the method proposed by~\cite{boracchi2012modeling} and~\cite{Kupyn2018deblurgan}, where random blur trajectories are generated and convolved with sharp images to obtain their corresponding blurred counterparts. The position of the next point in the trajectory is stochastically determined based on the velocity and location of the preceding point, incorporating Gaussian noise and an inertial effect influenced by both deterministic and impulsive factors.  

To better exploit the high-quality image generation capabilities of diffusion models and mitigate the limitations of existing datasets, we utilize the high-resolution DF2K dataset, which consists of DIV2K~\cite{Agustsson_2017_CVPR_Workshops} and Flickr2K~\cite{Lim_2017_CVPR_Workshops}, totaling 3,550 high-resolution images. We crop these images into 512 × 512 patches with overlap, resulting in 121K patches. Subsequently, we apply the randomly generated blur trajectories to these patches, generating the synthetic blurred dataset. Some samples are exhibited in~\cref{fig: syn}.

\vspace{-2mm}
\section{Experiments}
\label{sec: experiments}
\vspace{-0.5mm}
\subsection{Experiment Settings}
\vspace{-0.5mm}

\noindent \textbf{Data and Evaluation.} Following prior approaches in image deblurring, we assess our method using widely used GoPro~\cite{Nah2017deep} and RealBlur~\cite{Rim2020real} dataset. The GoPro dataset comprises 2,103 pairs of sharp and blurry images for training, along with 1,111 pairs for testing.  
The RealBlur dataset is divided into two subsets: RealBlur-J and RealBlur-R, each containing 3,758 training pairs and 980 testing pairs. We train our model using the GoPro training set and evaluate it on both the GoPro test set and, in a zero-shot manner, on the RealBlur-J and RealBlur-R test sets. For full-reference metics, we adopt LPIPS~\cite{lpips} and DISTS~\cite{dists}. For no-reference metrics, we utilize CLIPIQA~\cite{clipiqa}, NIQE~\cite{niqe}, MUSIQ~\cite{musiq}, and MANIQA~\cite{maniqa}.

\noindent \textbf{Implementation Details.}
Our model training consists of two stages. The first stage has been described in detail in~\cref{sec: vaeft}, where all training images have a fixed resolution of 512. This stage lasts for a total of 15K iterations.  

In the second stage (\cref{sec: ldmtrain}), we adopt a progressive learning strategy, where the image resolution is gradually increased as [160, 192, 256, 320, 384, 448, 512], with each resolution trained for 10K iterations. To prevent excessive blurriness in the restored images, we set \(\lambda_1\) and \(\lambda_2\) in~\cref{ldmloss} to 1 for 100K iterations. We use PyTorch to implement with 4 NVIDIA A800-80GB GPUs.

\vspace{-2.mm}
\subsection{Evaluations}
\vspace{-1.5mm}
We compare our OSDD with various models in Tabs.~\ref{tab: iqa_4} and~\ref{tab: bigtab_dists}, like MPRNet~\cite{Zamir2021multi}, HINet~\cite{Chen2021hinet}, Restormer~\cite{Zamir2021restormer}, Uformer~\cite{Wang2022uformer}, NAFNet~\cite{Chen2022simple}, Stripformer~\cite{Tsai2022Stripformer}, HI-Diff~\cite{hi-diff}, UFPNet~\cite{fang2023UFPNet}, DiffIR~\cite{Xia_2023_ICCV} and AdaRevD-L~\cite{xintm2024AdaRevD}. We also train an OSEDiff~\cite{wu2024one} model on GoPro.

\begin{table}[t]
    \centering
    \resizebox{1\linewidth}{!}{%
    \setlength{\tabcolsep}{1.5mm}
    \begin{tabular}{l|cc|cc|cc} 
    \toprule
    \rowcolor{row_gray} & \multicolumn{2}{c|}{GoPro} & \multicolumn{2}{c|}{RealBlur-J} & \multicolumn{2}{c}{RealBlur-R} \\ 
\rowcolor{row_gray} \multirow{-2}{*}{Method} & LPIPS & DISTS & LPIPS & DISTS & LPIPS & DISTS \\ \midrule
       MPRNet~\cite{Zamir2021multi} & 0.0886 & 0.0749 & 0.1587 & 0.1142 & 0.0761 & 0.1012 \\
       HINet~\cite{Chen2021hinet} & 0.0880 & 0.0703 & 0.1740 & 0.1228 & 0.0777 & 0.1032\\
       Restormer~\cite{Zamir2021restormer} & 0.0841 & 0.0724 & 0.1561 & 0.1112 & \textcolor{tab_blue}{0.0608} & 0.0833\\
       Uformer~\cite{Wang2022uformer} & 0.0868 & 0.0725 & 0.1474 & 0.1043 & 0.0615 & \textcolor{tab_blue}{0.0821}\\
       NAFNet~\cite{Chen2022simple} & 0.0781 & 0.0678 & 0.1688 & 0.1174 & 0.0683 & 0.0865\\
       Stripformer~\cite{Tsai2022Stripformer}& 0.0771 & 0.0680 & - & - & - & -\\
       UFPNet~\cite{fang2023UFPNet} & \textcolor{tab_blue}{0.0764} & 0.0666 & 0.1441 & 0.1101 & 0.0615 & 0.0853\\
       AdaRevD-L~\cite{xintm2024AdaRevD} & \textcolor{tab_red}{0.0712} & 0.0672 & 0.1408 & 0.1037 & 0.0621 & 0.0846 \\ 
       \midrule
       HI-Diff~\cite{hi-diff} & 0.0799 & 0.0710 & 0.1470 & 0.1050 & \textcolor{tab_red}{0.0602} & \textcolor{tab_red}{0.0814} \\
       DiffIR~\cite{Xia_2023_ICCV} & 0.0787 & 0.0702 & 0.1541 & 0.1079 & 0.0633 & 0.0845 \\
       OSEDiff~\cite{wu2024one} & 0.1719 & 0.0825 & 0.1793 & 0.1198 & 0.1057 & 0.1322 \\ \midrule
       OSDD ($\gamma=1$) & 0.1002 & \textcolor{tab_red}{0.0536} & \textcolor{tab_red}{0.1177} & \textcolor{tab_blue}{0.0859} & 0.0659 & 0.0893 \\
       OSDD ($\gamma=0.7$) & 0.1211 & \textcolor{tab_blue}{0.0606} & \textcolor{tab_blue}{0.1178} & \textcolor{tab_red}{0.0831} & 0.0683 & 0.0925 \\
       \bottomrule
    \end{tabular}}
    \vspace{-3.mm}
    \caption{Comparisons on GoPro~\cite{Nah2017deep} and RealBlur~\cite{Rim2020real} datasets. All models are trained only on GoPro.} 
    \label{tab: bigtab_dists}
    \vspace{-8.mm}
\end{table}

\begin{figure*}[t]
\scriptsize
\centering
\begin{tabular}{cc}
\hspace{-0.55cm}
\begin{adjustbox}{valign=t}
\begin{tabular}{c}
\includegraphics[width=0.1420\textwidth,height=0.09\textwidth]{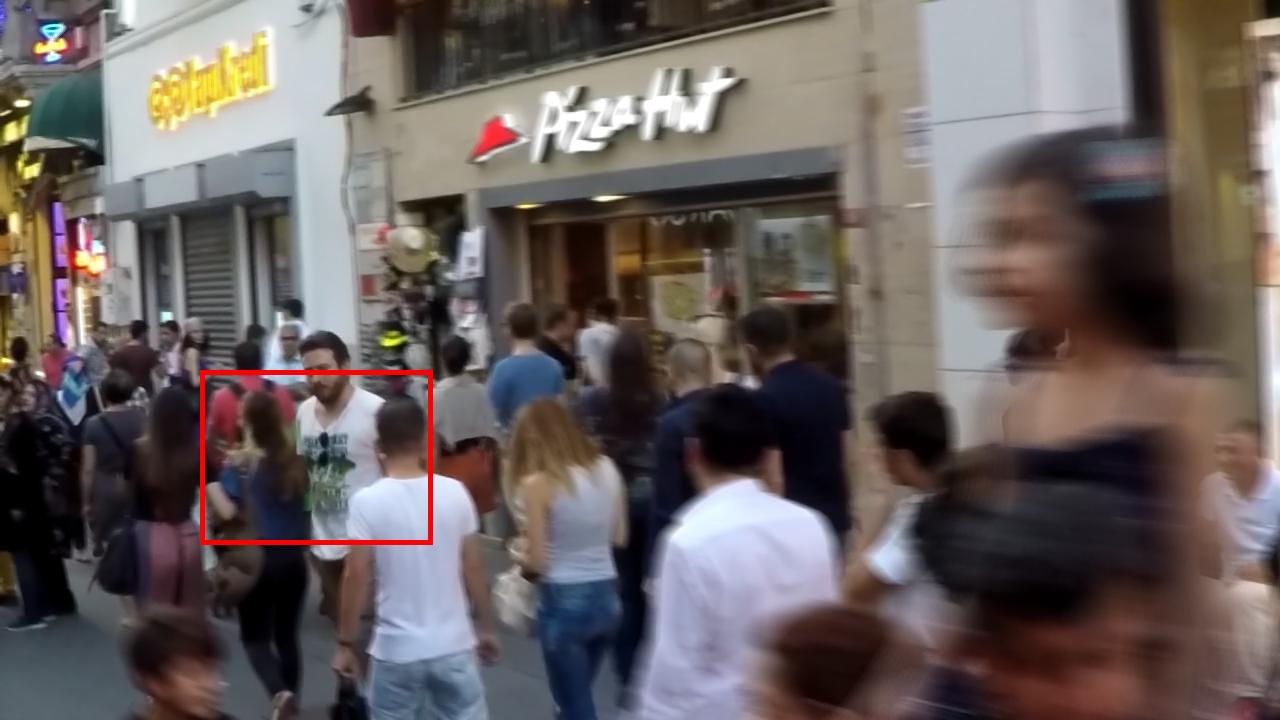}
\\
GOPRO384-004003
\end{tabular}
\end{adjustbox}
\hspace{-0.48cm}
\begin{adjustbox}{valign=t}
\begin{tabular}{ccccccc}
\includegraphics[width=0.12\textwidth, height=0.09\textwidth]{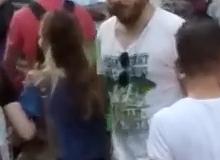} \hspace{-4.5mm} &
\includegraphics[width=0.12\textwidth, height=0.09\textwidth]{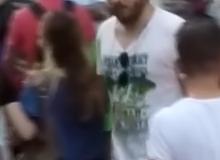} \hspace{-4.5mm} &
\includegraphics[width=0.12\textwidth, height=0.09\textwidth]{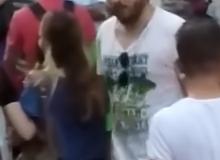} \hspace{-4.5mm} &
\includegraphics[width=0.12\textwidth, height=0.09\textwidth]{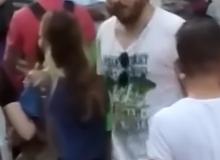} \hspace{-4.5mm} &
\includegraphics[width=0.12\textwidth, height=0.09\textwidth]{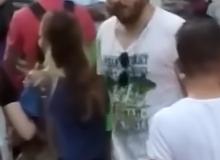} \hspace{-4.5mm} &
\includegraphics[width=0.12\textwidth, height=0.09\textwidth]{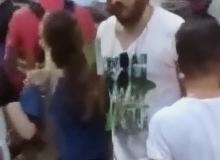} \hspace{-4.5mm} &
\includegraphics[width=0.12\textwidth, height=0.09\textwidth]{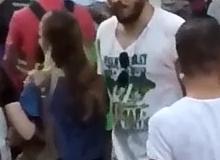} \hspace{-4.5mm}  
\\
HQ \hspace{-4.5mm} &
Blurry \hspace{-4.5mm}  &
UFPNet ~\cite{fang2023UFPNet} \hspace{-4.5mm} &
HI-Diff~\cite{hi-diff} \hspace{-4.5mm}  &
AdaRevD~\cite{xintm2024AdaRevD} \hspace{-4.5mm} &
OSEDiff~\cite{wu2024one} \hspace{-4.5mm} &
OSDD (ours) \hspace{-4.5mm}
\\
\end{tabular}
\end{adjustbox}

\end{tabular} \\

\begin{tabular}{cc}
\hspace{-0.55cm}
\begin{adjustbox}{valign=t}
\begin{tabular}{c}
\includegraphics[width=0.1420\textwidth,height=0.09\textwidth]{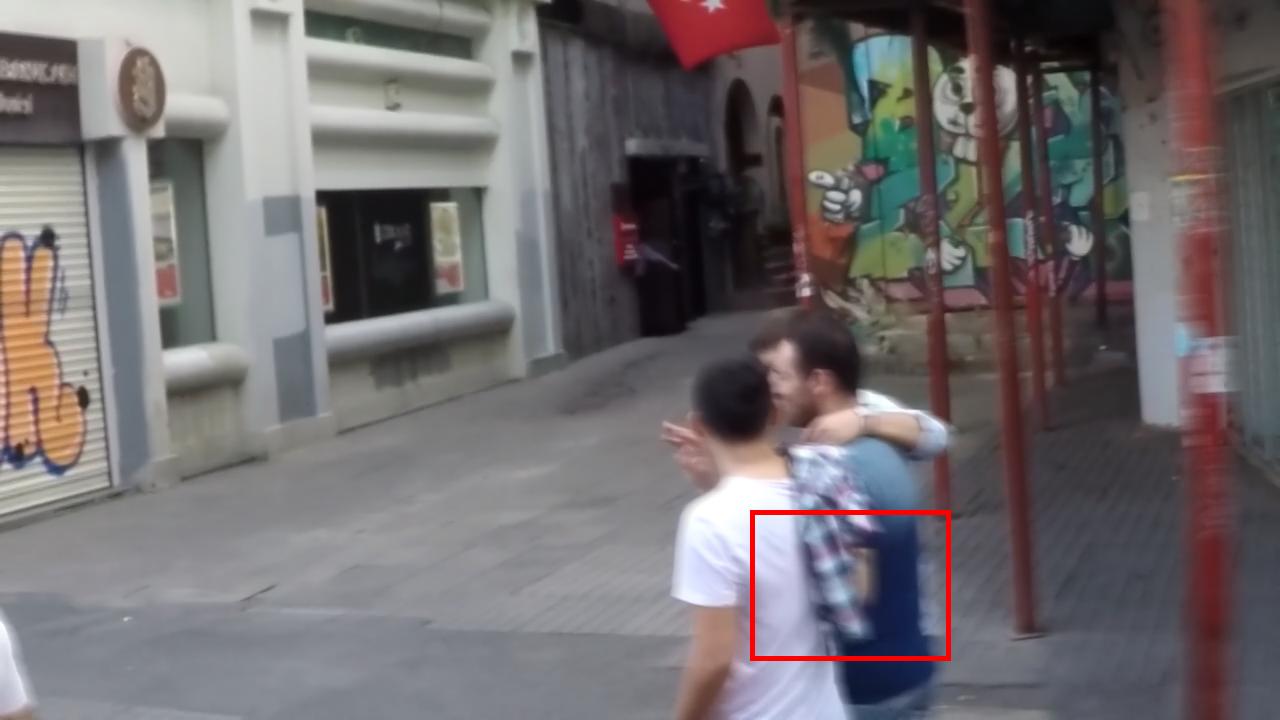}
\\
GOPRO385-003047
\end{tabular}
\end{adjustbox}
\hspace{-0.48cm}
\begin{adjustbox}{valign=t}
\begin{tabular}{ccccccc}
\includegraphics[width=0.12\textwidth, height=0.09\textwidth]{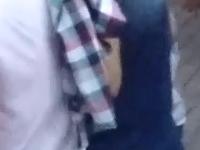} \hspace{-4.5mm} &
\includegraphics[width=0.12\textwidth, height=0.09\textwidth]{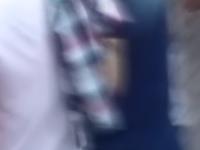} \hspace{-4.5mm} &
\includegraphics[width=0.12\textwidth, height=0.09\textwidth]{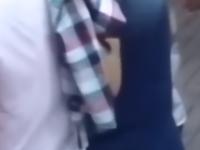} \hspace{-4.5mm} &
\includegraphics[width=0.12\textwidth, height=0.09\textwidth]{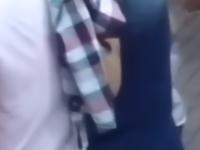} \hspace{-4.5mm} &
\includegraphics[width=0.12\textwidth, height=0.09\textwidth]{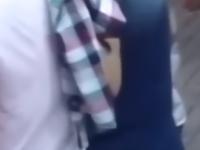} \hspace{-4.5mm} &
\includegraphics[width=0.12\textwidth, height=0.09\textwidth]{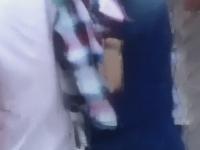} \hspace{-4.5mm} &
\includegraphics[width=0.12\textwidth, height=0.09\textwidth]{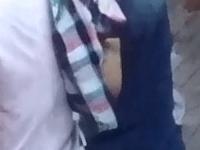} \hspace{-4.5mm}  
\\
HQ \hspace{-4.5mm} &
Blurry \hspace{-4.5mm}  &
UFPNet ~\cite{fang2023UFPNet} \hspace{-4.5mm} &
HI-Diff~\cite{hi-diff} \hspace{-4.5mm}  &
AdaRevD~\cite{xintm2024AdaRevD} \hspace{-4.5mm} &
OSEDiff~\cite{wu2024one} \hspace{-4.5mm} &
OSDD (ours) \hspace{-4.5mm}
\\
\end{tabular}
\end{adjustbox}

\end{tabular} \\

\begin{tabular}{cc}
\hspace{-0.55cm}
\begin{adjustbox}{valign=t}
\begin{tabular}{c}
\includegraphics[width=0.1420\textwidth,height=0.09\textwidth]{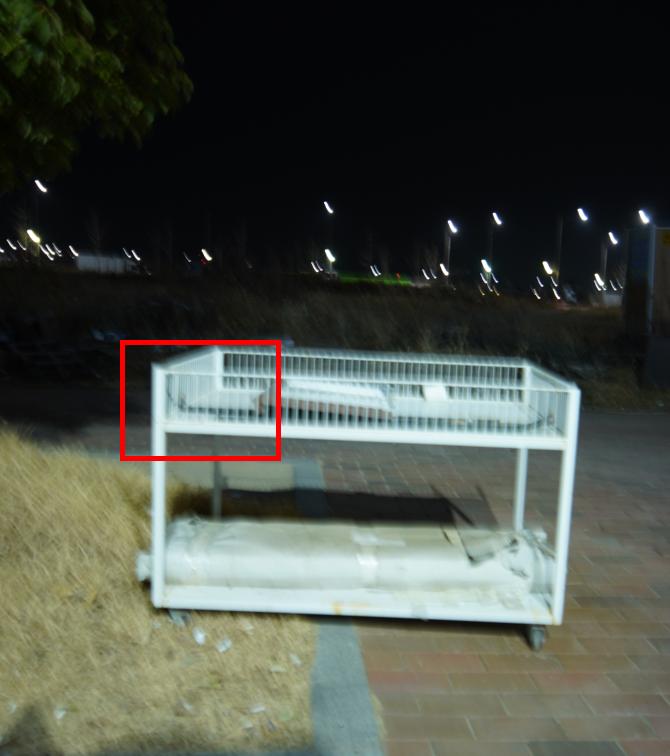}
\\
RealBlur-J scene120-9
\end{tabular}
\end{adjustbox}
\hspace{-0.48cm}
\begin{adjustbox}{valign=t}
\begin{tabular}{ccccccc}
\includegraphics[width=0.12\textwidth, height=0.09\textwidth]{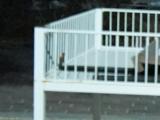} \hspace{-4.5mm} &
\includegraphics[width=0.12\textwidth, height=0.09\textwidth]{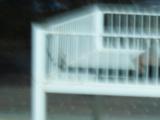} \hspace{-4.5mm} &
\includegraphics[width=0.12\textwidth, height=0.09\textwidth]{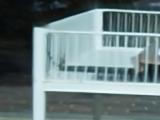} \hspace{-4.5mm} &
\includegraphics[width=0.12\textwidth, height=0.09\textwidth]{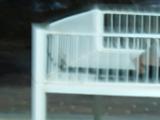} \hspace{-4.5mm} &
\includegraphics[width=0.12\textwidth, height=0.09\textwidth]{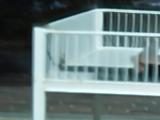} \hspace{-4.5mm} &
\includegraphics[width=0.12\textwidth, height=0.09\textwidth]{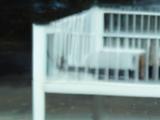} \hspace{-4.5mm} &
\includegraphics[width=0.12\textwidth, height=0.09\textwidth]{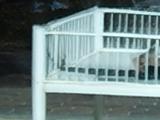} \hspace{-4.5mm}  
\\
HQ \hspace{-4.5mm} &
Blurry \hspace{-4.5mm}  &
UFPNet ~\cite{fang2023UFPNet} \hspace{-4.5mm} &
HI-Diff~\cite{hi-diff} \hspace{-4.5mm}  &
AdaRevD~\cite{xintm2024AdaRevD} \hspace{-4.5mm} &
OSEDiff~\cite{wu2024one} \hspace{-4.5mm} &
OSDD (ours) \hspace{-4.5mm}
\\
\end{tabular}
\end{adjustbox}

\end{tabular} \\

\begin{tabular}{cc}
\hspace{-0.55cm}
\begin{adjustbox}{valign=t}
\begin{tabular}{c}
\includegraphics[width=0.1420\textwidth,height=0.090\textwidth]{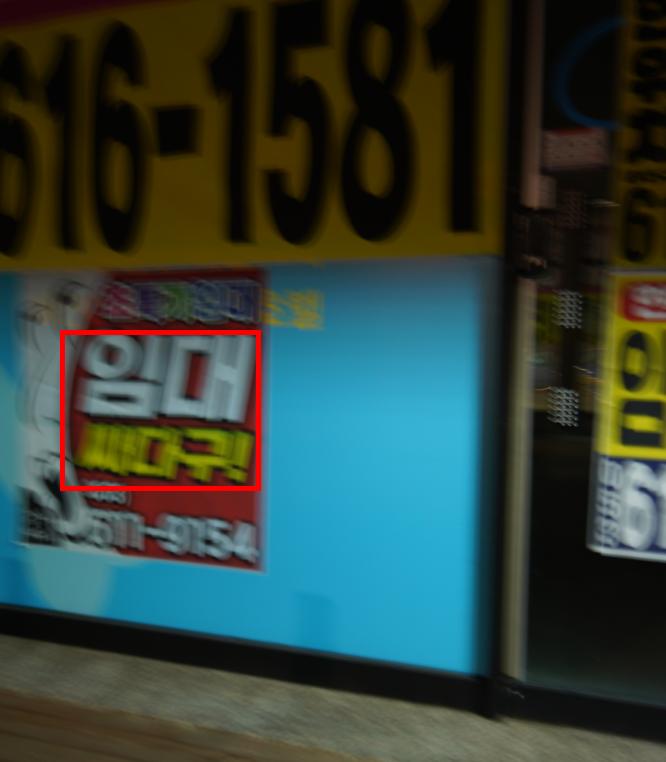}
\\
RealBlur-J scene142-11
\end{tabular}
\end{adjustbox}
\hspace{-0.48cm}
\begin{adjustbox}{valign=t}
\begin{tabular}{ccccccc}
\includegraphics[width=0.12\textwidth, height=0.09\textwidth]{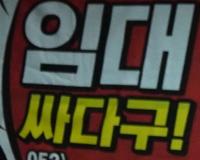} \hspace{-4.5mm} &
\includegraphics[width=0.12\textwidth, height=0.09\textwidth]{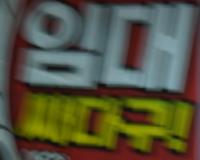} \hspace{-4.5mm} &
\includegraphics[width=0.12\textwidth, height=0.09\textwidth]{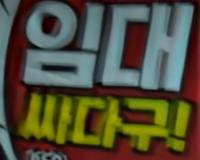} \hspace{-4.5mm} & 
\includegraphics[width=0.12\textwidth, height=0.09\textwidth]{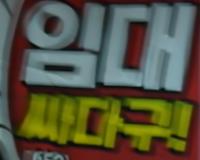} \hspace{-4.5mm} &
\includegraphics[width=0.12\textwidth, height=0.09\textwidth]{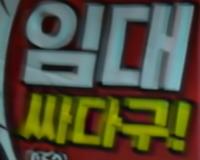} \hspace{-4.5mm} &
\includegraphics[width=0.12\textwidth, height=0.09\textwidth]{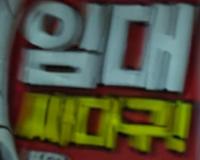} \hspace{-4.5mm} &
\includegraphics[width=0.12\textwidth, height=0.09\textwidth]{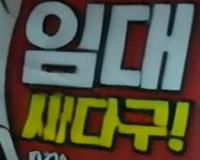} \hspace{-4.5mm}
\\
HQ \hspace{-4.5mm} &
Blurry \hspace{-4.5mm}  &
UFPNet ~\cite{fang2023UFPNet} \hspace{-4.5mm} &
HI-Diff~\cite{hi-diff} \hspace{-4.5mm}  &
AdaRevD~\cite{xintm2024AdaRevD} \hspace{-4.5mm} &
OSEDiff~\cite{wu2024one} \hspace{-4.5mm} &
OSDD (ours) \hspace{-4.5mm}
\\
\end{tabular}
\end{adjustbox}

\end{tabular}

\vspace{-3mm}
\caption{Visual results on GoPro and RealBlur-J. Our model shows a distinct advantage in recovering fine details.}
\label{fig: vis}
\vspace{-5.5mm}
\end{figure*}

\noindent \textbf{Quantitative Results.}
We treat LPIPS and DISTS as metrics for model fidelity, while the remaining no-reference (NF) metrics as indicators of the perception quality and naturalness. As shown in~\cref{tab: iqa_4}, we almost beat all of the baselines on the four NF metrics. And in ~\cref{tab: bigtab_dists}, our model achieves state-of-the-art DISTS metric on GoPro, the best on LPIPS and DISTS on RealBlur-J, and decent performance on RealBlur-R. In~\cref{tab: iqa_4} and~\ref{tab: bigtab_dists}, our OSDD ($\gamma=1$) denote the model without $LoRA_s$ that trained on synthetic datasets; $\gamma=0.7$ means we leverage 70\% on the GoPro-trained $LoRA_r$ module while 30\% on the $LoRA_s$.

In~\cref{tab: iqa_4}, OSDD ($\gamma=1$) exhibits significant \textbf{perceptual collapse}, with lower scores on NF metrics like CLIPIQA, MUSIQ, and MANIQA compared to other methods. This collapse is also evident in OSEDiff, which performs poorly in both metrics and visual quality, despite using a score distillation focused on reality perception. When the weight factor $\gamma$ is reduced, assigning some weight to the $LoRA_s$, the perceptual metrics generally improve significantly, with no or minimal degradation in LPIPS and DISTS. This highlights the importance of synthetic deblurring data pre-training and the effectiveness of the DDA.

According to~\cref{tab: bigtab_dists}, our model demonstrates strong generalization capabilities, achieving state-of-the-art (SOTA) on the RealBlur-J dataset. We attribute this robust generalization to the large pre-trained knowledge embedded in our OSDD framework, which enables it to go beyond simply modeling local blur patterns. However, our model does not achieve the best performance on the RealBlur-R dataset, primarily because the images in RealBlur-R were captured in extremely dark environments, with some images even lacking discernible object contours. This poses a challenge, as a well-trained diffusion model typically does not generate images with such extremely dark scenes.

\textbf{Speed.}
The single-step diffusion model significantly accelerates the image generation process. This advantage is demonstrated in~\cref{tab: speed}, where our model shows a noticeable improvement in inference speed.

\begin{table}[t]
    \centering
    \resizebox{1\linewidth}{!}{
    \setlength{\tabcolsep}{3.mm}
    \begin{tabular}{l|c|c|c|c} \toprule
      \rowcolor{row_gray}  Model & Restormer~\cite{Zamir2021restormer} & UFPNet~\cite{fang2023UFPNet} & AdaRevD-L~\cite{xintm2024AdaRevD} & OSDD \\
      \midrule
       GoPro & 1.20 & 1.47 & 2.05 & 0.88 \\
       RealBlur-J & 0.68 & 0.81 & 1.18 & 0.65 \\
       \bottomrule
    \end{tabular}}
    \vspace{-3mm}
    \caption{Inference speed (sec/image).}
    \label{tab: speed}
    \vspace{-6.5mm}
\end{table}

\textbf{Visualizations.}
Some visualizations are shown in~\cref{fig:first_comp} and~\cref{fig: vis}. OSDD performs better in many fine details and the visual results exhibit sharper contours and provide a superior visual experience. Besides, in the supplementary file, we also provide many samples that visually our OSDD ($\gamma=0.7$) outperforms OSDD ($\gamma=1$), where OSDD ($\gamma=0.7$) can more robustly handle various types of blur and generate images that are closer to real-world ones.

\vspace{-1.5mm}
\subsection{Ablation Study.}
\vspace{-1.5mm}

\begin{table}[t]
    \centering
    \resizebox{1\linewidth}{!}{%
    \setlength{\tabcolsep}{2.mm}
    \begin{tabular}{l|cc|cc|cc} 
    \toprule
   \rowcolor{row_gray} & \multicolumn{2}{c|}{GoPro} & \multicolumn{2}{c|}{RealBlur-J} & \multicolumn{2}{c}{RealBlur-R} \\ 
\rowcolor{row_gray} \multirow{-2}{*}{Method} & MAE$_{pixel}$ & LPIPS & MAE$_{pixel}$ & LPIPS & MAE$_{pixel}$ & LPIPS \\ \midrule
    VAE w/o ft & 4.85 & 0.0539 & 3.26 & 0.0367 & 1.56 & 0.0205 \\
    VAE w/ ft & 3.98 & 0.0356 & 3.14 & 0.0373 & 1.17 & 0.0244 \\ 
    eVAE$_2$ & 3.72 & 0.0335 & 3.29 & 0.0455 & 1.22 & 0.0219 \\ 
    eVAE$_3$ & 3.34 & 0.0280 & 3.03 & 0.0432 & 1.22 & 0.0228 \\ 
    \bottomrule
    \end{tabular}}
    \vspace{-3.mm}
    \caption{VAE enhancement. VAE w/ ft refers to directly fine-tuning the original VAE with HQs, without enhancement. eVAE$_2$ represents a variant where Encoder 2 is frozen, training only Encoder 1 and the Decoder. eVAE$_3$ indicates the configuration where Encoder 1, Encoder 2, and the Decoder are trained jointly.}
    \vspace{-4.mm}
    \label{tab: ftvae}
\end{table}

\begin{figure}[t]
\scriptsize
\centering
\begin{tabular}{ccc}
\hspace{-0.45cm}
\begin{adjustbox}{valign=t}
\begin{tabular}{c}
\includegraphics[width=0.25\textwidth,height=0.25\textwidth]{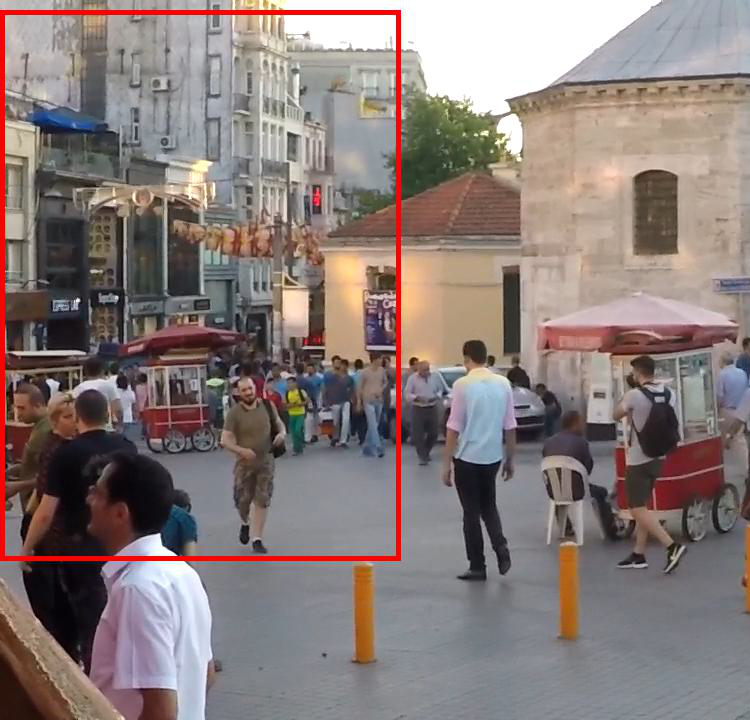}
\\
GOPRO384\_11\_00-000010
\end{tabular}
\end{adjustbox}
\hspace{-0.46cm}
\begin{adjustbox}{valign=t}
\begin{tabular}{cccc}
\includegraphics[width=0.081\textwidth]{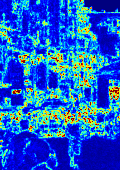} \hspace{-4mm} &
\includegraphics[width=0.081\textwidth]{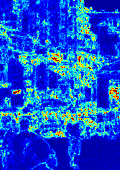} \hspace{-4mm} &
\\
VAE w/o ft \hspace{-4mm}
 &
VAE w/ ft \hspace{-4mm}  &
\\
\includegraphics[width=0.081\textwidth]{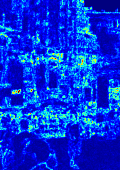} \hspace{-4mm} &
\includegraphics[width=0.081\textwidth]{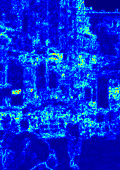} \hspace{-4mm} &
\\ 
eVAE$_2$ \hspace{-4mm}  &
eVAE$_3$ \hspace{-4mm}
\\
\end{tabular}
\end{adjustbox}
\hspace{-0.46cm}
\begin{adjustbox}{valign=t}
\begin{tabular}{c}
\includegraphics[width=0.0390\textwidth,height=0.250\textwidth]{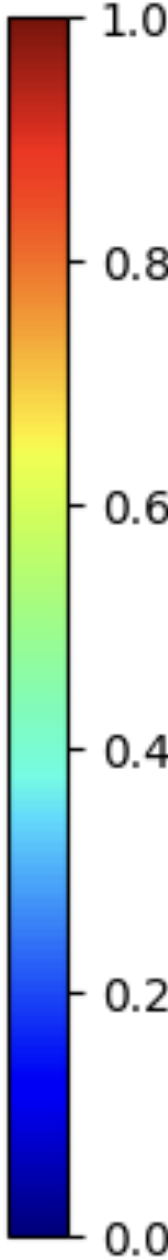}
\end{tabular}
\end{adjustbox}

\end{tabular}
\vspace{-3.5mm}
\caption{Different VAE reconstruction errors. }
\label{fig: vae_cmp}
\vspace{-8.mm}
\end{figure}

\textbf{VAE enhancement.}
We fine-tuned the first-stage eVAE on the GoPro dataset and tested its performance on both GoPro and RealBlur. The results are obtained by feeding the HQ images into eVAE and measuring the similarity between the input and the reconstructions, which is the upper bound of our OSDD framework. The results in~\cref{tab: ftvae} indicate that our proposed improvements significantly reduce pixel-level errors (scaled to $0\sim255$) and LPIPS scores on the GoPro dataset, while also showing great generalization to RealBlur. This enhancement elevates the overall performance upper bound of the task, playing a pivotal role in the second-stage training process. We visualize a comparison sample in~\cref{fig: vae_cmp}, which directly shows the reconstruction error differences. We also trained a model without eVAE, showing clear degradation (shown in the supplementary file).

\begin{figure}[t]
    \centering
    \includegraphics[width=1.0\linewidth]{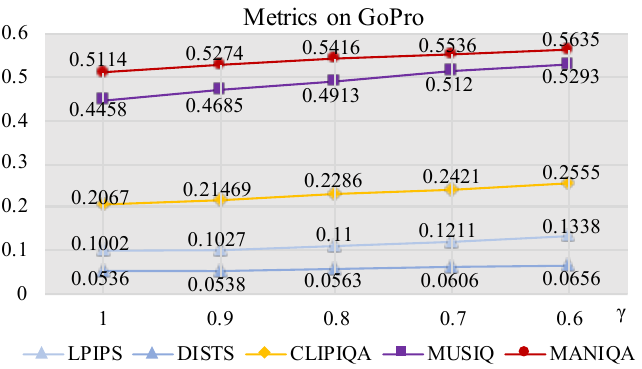}
    \vspace{-6mm}
    \caption{On GoPro, reducing $\gamma$ in OSDD leads to remarkably improved visual perception while incurring minimal loss in fidelity.}
    \label{fig: gamma_gopro}
    \vspace{-3.mm}
\end{figure}

\begin{figure}[t]
    \centering
    \includegraphics[width=1.0\linewidth]{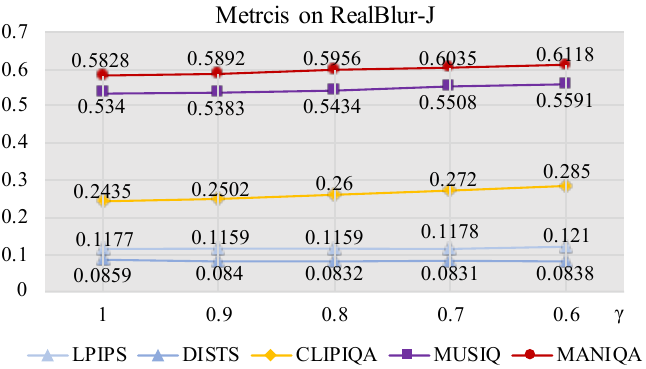}
    \vspace{-7mm}
    \caption{On RealBlur-J, reducing the weight of $\gamma$ in OSDD leads to improved visual perception without compromising fidelity.}
    \label{fig: gamma-realj}
    \vspace{-6mm}
\end{figure}

\textbf{Dynamic Dual-Adapter.}
We progressively reduce $\gamma$ of the Dynamic Dual-Adapter (DDA) during the inference phase and illustrate the trends of full-reference (FR) metrics, including LPIPS and DISTS, as well as three NR metrics, CLIPIQA, MUSIQ, and MANIQA, in Figs.~\ref{fig: gamma_gopro} and~\ref{fig: gamma-realj}. The MUSIQ metric is normalized to the range of 0 to 1. It can be observed that by decreasing the weight $\gamma$, our OSDD significantly improves NR metrics while almost fully preserving FR metrics. This demonstrates the effectiveness of our DDA design and the benefits of synthetic data, which can mitigate perceptual collapse, enhance perceptual quality, and maintain fidelity. To better balance the perception-fidelity tradeoff, we carefully set $\gamma$ as \textbf{0.7}, maximizing the perception quality (NF metrics) while keeping the fidelity loss (LPIPS, DISTS) within an acceptable range. 

\textbf{Synthetic Data Pre-training.} We also conduct an experiment where only the weight of $LoRA_r$ was reduced while $LoRA_s$ remained unused, \ie, adjusting $\gamma$ in~\cref{eq: lora} while keeping the coefficient of $LoRA_s$ at zero. The results, presented in~\cref{tab: wodf}, show the performance degradation without leveraging synthetic data, demonstrate the importance of $LoRA_s$, and suggest that maintaining the weighted sum at 1 is preferable. We also conducted an experiment that the model trained only on synthetic data will generate high-quality images but perform poorly on GoPro and RealBlur. More details are in the supplementary file.

\textbf{Other Training Schedule.}
Initially, we employed a single LoRA module, first training on our high-quality synthetic data and then on the GoPro dataset. The intention was to leverage the synthetic data fine-tuning to achieve a more robust parameter initialization. However, as shown in~\cref{tab: dfinit}, this approach (denoted as OSDD$^*$ in the table) does not outperform our DDA-designed OSDD, and the results are similar to OSDD ($\gamma=1$), showing limited perceptual quality. This suggests that initialization alone may not significantly contribute to guiding the model toward a better convergence state. Therefore, it is more effective to retain the weights and utilize them as DDA.

\begin{table}[t]
    \centering
    \resizebox{1.\linewidth}{!}{%
    \setlength{\tabcolsep}{0.5mm}
    \begin{tabular}{l|cc|cccc} 
    \toprule
  \rowcolor{row_gray} GoPro & LPIPS$\downarrow$ & DISTS$\downarrow$ & CLIPIQA$ \uparrow $ & NIQE$ \downarrow $ & MUSIQ$ \uparrow $ & MANIQA$ \uparrow$ \\ \midrule
    $\gamma=1$ & 0.1002 & 0.0536 & 0.2067 & 4.23 & 44.58 & 0.5114 \\ 
       \midrule
    $\gamma=0.8$ & 0.1100 & 0.0563 & 0.2286 & 4.13 & 49.13 & 0.5416 \\ 
    $\gamma=0.8$ w/o $LoRA_s$ & 0.1231 & 0.0685 & 0.2189 & 4.13 & 49.23 & 0.5250 \\ 
       \midrule
    $\gamma=0.6$ & 0.1338 & 0.0656 & 0.2555 & 4.10 & 52.93 & 0.5635 \\ 
    $\gamma=0.6$ w/o $LoRA_s$ & 0.2178 & 0.1201 & 0.2189 & 4.47 & 45.87 & 0.4870 \\ 
    \midrule
    \midrule
   \rowcolor{row_gray} RealBlur-J & LPIPS $\downarrow$ & DISTS$\downarrow$ & CLIPIQA$ \uparrow $ & NIQE$ \downarrow $ & MUSIQ$ \uparrow $ & MANIQA$ \uparrow$ \\ \midrule
    $\gamma=1$ & 0.1177 & 0.0859 & 0.2435 & 4.57 & 53.40 & 0.5828 \\ 
       \midrule
    $\gamma=0.8$ & 0.1159 & 0.0832 & 0.2600 & 4.54 & 54.34 & 0.5956 \\ 
    $\gamma=0.8$ w/o $LoRA_s$ & 0.1308 & 0.9050 & 0.2349 & 4.49 & 53.56 & 0.5784 \\ 
       \midrule
    $\gamma=0.6$ & 0.1210 & 0.0838 & 0.2850 & 4.53 & 55.91 & 0.6118 \\ 
    $\gamma=0.6$ w/o $LoRA_s$ & 0.1884 & 0.1249 & 0.2212 & 4.71 & 52.01 & 0.5538 \\ 
       \bottomrule
    \end{tabular}}
    \vspace{-3mm}
    \caption{Synthetic data pre-training experiments for OSDD. Leveraging weights on $LoRA_s$ is important. Removing $LoRA_s$ and simply lowering $\gamma$ of $LoRA_r$ fails to boost the performance.}
    \vspace{-4mm}
    \label{tab: wodf}
\end{table}

\begin{table}[t]
    \centering
    \resizebox{1.\linewidth}{!}{%
    \setlength{\tabcolsep}{0.5mm}
    \begin{tabular}{l|cc|cccc} 
    \toprule
  \rowcolor{row_gray} GoPro & LPIPS$\downarrow$ & DISTS$\downarrow$ & CLIPIQA$ \uparrow $ & NIQE$ \downarrow $ & MUSIQ$ \uparrow $ & MANIQA$ \uparrow$ \\ \midrule
    OSDD ($\gamma=0.8$) & 0.1100 & 0.0563 & 0.2286 & 4.13 & 49.13 & 0.5416 \\ 
    OSDD ($\gamma=1$) & 0.1002 & 0.0536 & 0.2067 & 4.23 & 44.58 & 0.5114 \\
    OSDD$^*$ & 0.1147 & 0.0542 & 0.2112 & 4.16 & 44.38 & 0.5156 \\ 
    \midrule
    \midrule
   \rowcolor{row_gray} RealBlur-J & LPIPS $\downarrow$ & DISTS$\downarrow$ & CLIPIQA$ \uparrow $ & NIQE$ \downarrow $ & MUSIQ$ \uparrow $ & MANIQA$ \uparrow$ \\ \midrule
    OSDD ($\gamma=0.8$) & 0.1159 & 0.0832 & 0.2600 & 4.54 & 54.34 & 0.5956 \\ 
    OSDD ($\gamma=1$) & 0.1177 & 0.0859 & 0.2435 & 4.57 & 53.40 & 0.5828 \\
    OSDD$^*$ & 0.1299 & 0.0912 & 0.2090 & 4.39 & 48.84 & 0.5517 \\ 
    \bottomrule
    \end{tabular}}
    \vspace{-3mm}
    \caption{Robust parameter initialization. Using a single LoRA module trained on synthetic data followed by fine-tuning on GoPro yields inferior results compared to our DDA design.}
    \label{tab: dfinit}
    \vspace{-6mm}
\end{table}

\vspace{-2mm}
\section{Conclusion}
\vspace{-1.5mm}
In this paper, we explore the application of diffusion models for single-image deblurring and propose a one-step diffusion model for efficient deblurring. To achieve fine-grained detail reconstruction, we introduce an enhanced variational autoencoder to address structural and pixel-level loss in diffusion models. Additionally, we construct a high-quality synthetic deblurring dataset to mitigate perceptual collapse and design a dynamic dual-adapter, which adaptively fuses pre-trained knowledge to enhance the perception quality while preserving fidelity. Extensive experiments have demonstrated that our proposed framework is more robust, generalizes well, and effectively handles more complex blurry patterns. Our work highlights the potential of diffusion models in deblurring and provides a new direction for efficient, high-quality image restoration.

{
    \small
    \bibliographystyle{ieeenat_fullname}
    \bibliography{arxiv-main}
}

\end{document}